\documentclass[letterpaper, 10 pt, conference]{ieeeconf}  %

\IEEEoverridecommandlockouts                              %
\IEEEoverridecommandlockouts                              %

\newcommand{\model}{\textsc{magic}\xspace}
\newcommand{\modelfull}{manipulation analogies for generalizable intelligent contacts\xspace}
\newcommand{\Modelfull}{Manipulation Analogies for Generalizable Intelligent Contacts\xspace}
\newcommand{\ours}{\model}

\usepackage{amsmath,amsfonts,bm}

\def\eqref#1{equation~\ref{#1}}

\def\1{\bm{1}}

\DeclareMathAlphabet{\mathsfit}{\encodingdefault}{\sfdefault}{m}{sl}
\SetMathAlphabet{\mathsfit}{bold}{\encodingdefault}{\sfdefault}{bx}{n}

\usepackage{color,xcolor}
\usepackage{epsfig}
\usepackage{graphicx}

\usepackage{adjustbox}
\usepackage{array}
\usepackage{booktabs}
\usepackage{colortbl}
\usepackage{wrapfig}
\usepackage{hhline}
\usepackage{multirow}
\usepackage{subcaption} %
\usepackage[skip=1pt,labelfont=bf,font=footnotesize]{caption}
\usepackage{wrapfig}
\usepackage{floatflt}

\usepackage{amsmath,amsfonts,amssymb,amsthm}
\usepackage{mathtools}  %
\usepackage{bm}
\usepackage{nicefrac}
\usepackage{microtype}
\usepackage{times} %
\usepackage{mathptmx} %

\usepackage{changepage}
\usepackage{extramarks}
\usepackage{fancyhdr}
\usepackage{lastpage}
\usepackage{setspace}
\usepackage{soul}
\usepackage{xspace}

\usepackage[pagebackref=true,breaklinks=true,colorlinks,citecolor=gray]{hyperref}

\usepackage{url}
\definecolor{urlblue}{rgb}{0,0,0.5}
\hypersetup{urlcolor=urlblue}
\usepackage[noadjust]{cite}

\usepackage{enumerate}
\usepackage{todonotes} %
\let\labelindent\relax
\usepackage{enumitem}  %

\usepackage{titlesec}

\usepackage{makecell}

\usepackage{pifont} %

\usepackage{algorithm}
\usepackage[noend]{algpseudocode}

\usepackage{framed}
\definecolor{shadecolor}{gray}{0.95}

\usepackage{xparse}
\usepackage{etoolbox}

\usepackage{courier}

\newcolumntype{L}[1]{>{\raggedright\let\newline\\\arraybackslash\hspace{0pt}}m{#1}}
\newcolumntype{C}[1]{>{\centering\let\newline\\\arraybackslash\hspace{0pt}}m{#1}}
\newcolumntype{R}[1]{>{\raggedleft\let\newline\\\arraybackslash\hspace{0pt}}m{#1}}

\newcommand{\fig}[1]{Fig.~\ref{#1}}
\newcommand{\tbl}[1]{Table~\ref{#1}}

\newcommand{\ignore}[1]{}

\makeatletter
\DeclareRobustCommand\onedot{\futurelet\@let@token\@onedot}
\def\@onedot{\ifx\@let@token.\else.\null\fi\xspace}

\def\eg{e.g\onedot} 
\def\ie{i.e\onedot} 
\def\cf{\emph{c.f}\onedot}

\makeatother

\definecolor{MyDarkBlue}{rgb}{0,0.08,1}
\definecolor{MyDarkGreen}{rgb}{0.02,0.6,0.02}
\definecolor{MyDarkRed}{rgb}{0.8,0.02,0.02}
\definecolor{MyDarkOrange}{rgb}{0.40,0.2,0.02}
\definecolor{MyPurple}{RGB}{111,0,255}
\definecolor{MyRed}{rgb}{1.0,0.0,0.0}
\definecolor{MyGold}{rgb}{0.75,0.6,0.12}
\definecolor{MyDarkgray}{rgb}{0.66, 0.66, 0.66}
\definecolor{JiayuanColor}{rgb}{0.60,0.43,0.48}

\newif\ifpropositionfirstitem
\propositionfirstitemtrue

\newcommand{\xhdr}[1]{\noindent\textbf{#1}}

\theoremstyle{plain}%

\title{\LARGE \bf
One-Shot Manipulation Strategy Learning\\by Making Contact Analogies
}

\author{Yuyao Liu$^{1,2*}$, Jiayuan Mao$^{1*}$, Joshua B. Tenenbaum$^{1}$, Tomás Lozano-Pérez$^{1}$, Leslie Pack Kaelbling$^{1}$ \\
$^{1}$Massachusetts Institute of Technology\\
$^{2}$Institute for Interdisciplinary Information Sciences, Tsinghua University\\
{\small \tt liuyuyao21@mails.tsinghua.edu.cn~\{jiayuanm,jbt\}@mit.edu~\{tlp,lpk\}@csail.mit.edu} \\
\thanks{* indicates equal contribution. Work done while Yuyao Liu was a visiting student at MIT.}%
\vspace{-1.5em}%
}

\begin{document}

\maketitle
\thispagestyle{empty}
\pagestyle{empty}

\begin{abstract}
We present a novel approach, \model (\modelfull), for one-shot learning of manipulation strategies with fast and extensive generalization to novel objects. By leveraging a reference action trajectory, \ours effectively identifies similar contact points and sequences of actions on novel objects to replicate a demonstrated strategy, such as using different hooks to retrieve distant objects of different shapes and sizes. Our method is based on a two-stage contact-point matching process that combines global shape matching using pretrained neural features with local curvature analysis to ensure precise and physically plausible contact points. We experiment with three tasks including scooping, hanging, and hooking objects. \ours demonstrates superior performance over existing methods, achieving significant improvements in runtime speed and generalization to different object categories. Website: \url{https://magic-2024.github.io/}.
\end{abstract}

\section{Introduction}\label{sec:intro}
\begin{figure*}[tp]
    \centering
    \includegraphics[width=\textwidth]{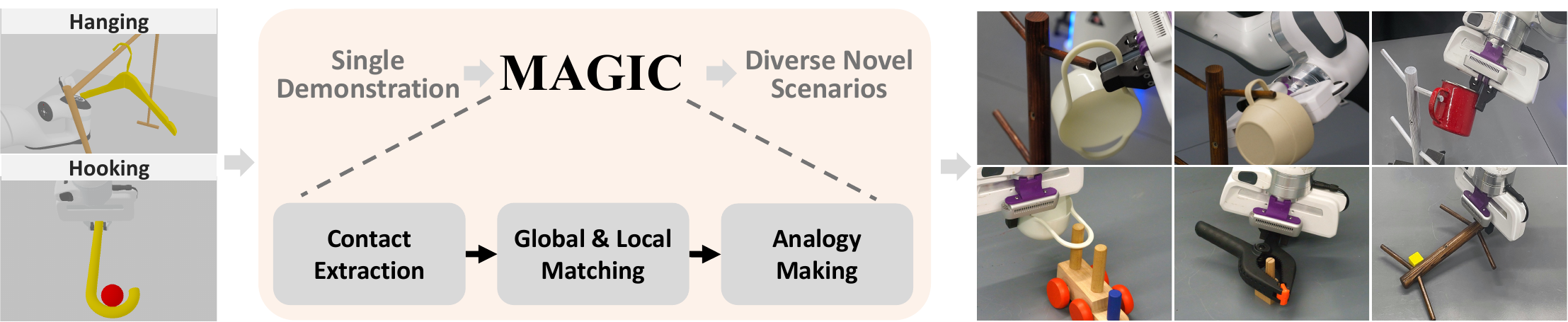}
    \vspace{-1em}
    \caption{We introduce \ours (\Modelfull), a pipeline that is capable of learning manipulation strategies from single demonstrations and applying them to novel objects.}
    \label{fig:teaser}
    \vspace{-2em}
\end{figure*}

A hallmark of human intelligence is flexible tool use: humans can quickly acquire new manipulation ``strategies'' from just a handful of demonstrations and apply these strategies across various scenarios, including generalization to novel objects of unseen categories. For example, as illustrated in \fig{fig:teaser}, even from a single demonstration of using a hook to reach distant objects or putting hangers on a rod, we can generalize to different object positions, sizes, and diverse categories, such as hangers and mugs.

Traditionally, two main approaches have been widely studied to build machines that can flexibly use tools: model-based and analytic approaches which take novel scenarios and goals and use built-in physical models to compute plans~\cite{cheng2021contact,cheng2022contact,pang2023global,mao2023learning}, and policy learning, which leverages various types of priors (\eg, object-based and part-based models) and pretrained neural features for generalization~\cite{simeonov2022neural,liu2023composable,huang2024copa,zhu2024orion}. However, both approaches have their limitations. Model-based planning generalizes well given accurate object and physical models. However, it is slow and usually does not benefit from learning. Policy learning approaches, on the other hand, are very efficient at performance time but usually exhibit limited generalization to novel objects and scenarios, particularly when the shape of the novel objects differs significantly from objects seen during training, such as generalizing from hangers to mugs.

In this paper, we present a novel approach, \ours (\modelfull), for one-shot manipulation strategy learning. Shown in \fig{fig:teaser}, given a {\it single} reference action trajectory (\eg, using a hook to reach for a distant object) and a novel scenario (\eg, with different tools and different objects), the goal of the algorithm is to generate a sequence of robot actions that apply a ``similar'' strategy to the test objects specified by users: in this example, having the target object moving along a certain direction for a given distance. \ours extends two critical insights into a broad class of manipulation strategies. First, many strategies such as hooking, hanging, hammering, pushing, reaching~\cite{qin2020keto, turpin2021gift}, stacking, pouring~\cite{lai2021functional, sieb20graph, xu2021affordance}, and cutting~\cite{xu2021affordance} can be characterized by a sequence of contact waypoints (\ie, the order in which contacts between objects and robot bodies are made); second, these contacts are characterized by forceful affordances between object pairs: a specific pair of contact points on two objects would enable the application of forces along certain directions. However, searching for contact points that would enable the specific affordance is generally challenging due to complex constraints on reachability, collision avoidance, and motion stability.

\ours tackles these challenges by combining data-driven and analytic approaches to generate contact waypoints in novel scenarios. In particular, it first extracts the sequence of contacts among objects in the reference trajectory and then proposes (pairs of) contact points that have similar global and local shape properties as the contact points in the reference, which can be used as guidance for motion planning or motion retargeting. Finally, it utilizes a physical simulator to discard trajectories that fail to achieve the goal due to collisions, unstable physical contact, or violations of joint and torque limits. Our key innovation lies in a novel global-to-local matching algorithm to find functional correspondences between the target objects and reference objects. Intuitively, a ``good'' contact point would satisfy both a global and a local matching property. First, the points on two objects should be on similar parts of the global shape (\eg, in the hook-using example, we need a contact point on the tool that is at the end of a long rod). Second, the hooking contact point should have a matched local curvature with the target object being hooked so that we can execute the actions stably. Therefore, we propose to use a pretrained visual feature-based correspondence matching to resolve the global matching property. This enables us to quickly search over different parts of the objects but the resulting contact point is usually not precise. Next, we use a local curvature-based matching algorithm to find the best contact point within a local region of the previously proposed contact point, which gives us precise and physically plausible (\eg, collision-free and physically stable) contacts.

Overall, \model tackles the problem of one-shot manipulation strategy learning by making analogies in contact waypoints.
We validate the effectiveness of our approach on three challenging tasks: scooping a ball against a concave arc with a spoon, hanging a mug onto a mug tree, and using tools to hook objects of varying sizes. Compared to global shape-matching algorithms, our framework achieves significant improvements when the reference objects are from different categories than the test objects. Compared to local shape-matching and simulation-based approaches, our framework is orders of magnitude faster --- for most test objects, we need to run simulations on fewer than three candidate contact points to find a solution. Finally, compared to pretrained feature-matching-based approaches, our method finds more precise and physically plausible solutions.

\section{Related Work}\label{sec:related_work}
\xhdr{Contact-Based Modeling in Robotic Manipulation.} Various approaches have been developed for planning manipulations in contact space, involving both rigid bodies and robotic hands \cite{trinkle1991framework,ji2001planning,yashima2003randomized,lee2015hierarchical,mordatch2012discovery,hauser2010multi,sleiman2019contact,aceituno2020global,you2021omnihang,huang2023autogenerated,wu2024oneshot}. Techniques such as CMGMP~\cite{xiao2001automatic,cheng2021contact,cheng2022contact} and the work of \cite{pang2023global} perform search over the exponential space of possible contact sequences to determine possible contact modes using three basic types: fixed, separating, and sliding contact, while \cite{mao2023learning} learns the types of the contact sequence from a single demonstration to guide the planning around them. By contrast, in this paper, we focus on leveraging global and local shape matching to generate analogical contact points using a single demonstration, which significantly improves the runtime efficiency of existing methods as well as the generalizations to unseen objects.

\xhdr{Affordance Learning and Keypoint Representations.} Object affordance characterizes how an object can interact with others, often represented through keypoints. Traditionally, this concept has been learned from large dataset ~\cite{gupta2007objects,zhu2015understanding,fang2020learning,qin2020keto,turpin2021gift,lai2021functional,mo2021where2act}. In contrast, our approach generates trajectories for novel objects from a single demonstration without requiring additional training data. More recent works~\cite{huang2023voxposer,ju2024robo,kuang2024ram,zhu2024densematcher,fangandliu2024moka} have explored one-shot or zero-shot affordance using vision and language foundation models; however, these methods operate primarily within the semantic domain and may not generalize well to creative, tool-using scenarios. Our work distinguishes itself by proposing precise, physically plausible contact points derived from pretrained visual features and advanced shape analysis algorithms.

\xhdr{Shape-Guided Tool Using.} For many years, people have attempted to harness shape information to guide tool usage through mathematical analysis \cite{asada1986curvature, jensen2002tool, brandi2014generalizing, rodriguez2018transferring, biza2023one} and data-driven approaches \cite{simeonov2022neural, thompson2021shape, manuelli2019kpam, gao2021kpam, turpin2021gift, wen2022you}. These methods often necessitate intricate human specifications or extensive in-domain datasets. In contrast, our approach integrates the off-the-shelf visual models, pretrained on general image datasets, with the generic geometrical property of curvature. This combination allows us to achieve effective and efficient tool-using guidance with minimal human intervention and without the need for large specialized datasets.

\section{One-Shot Manipulation Strategy Learning}\label{sec:approach}
\begin{figure*}[tp]
\centering
\includegraphics[width=\linewidth]{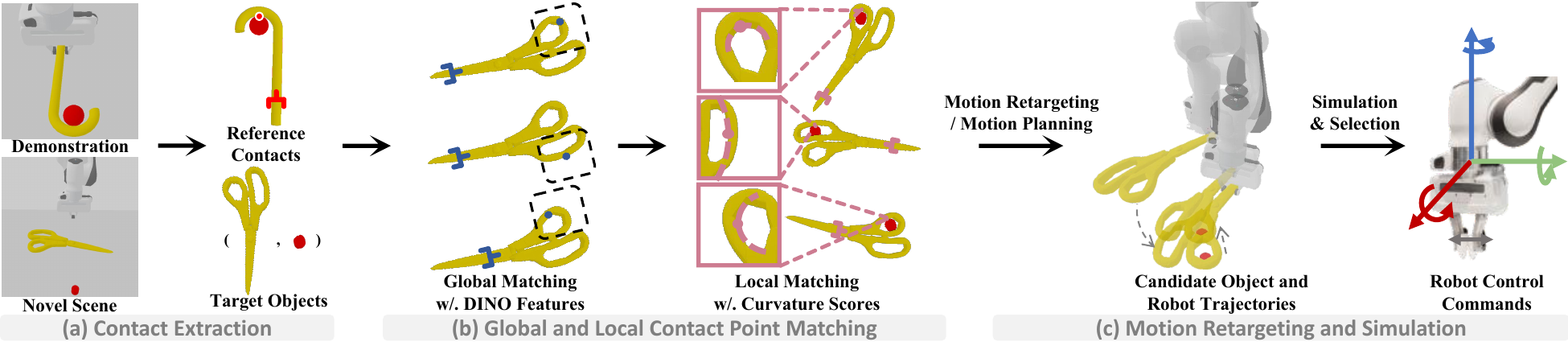}
\vspace{-1em}
\caption{The overall pipeline of \model. (a) We first extract contact points from the reference trajectory. (b) Then, we compute a global and local contact point matching score to select candidate contact points on novel objects. (c) The generated contact points will be used for motion retargeting or motion planning, and the final motion will be simulated and verified by a physical simulator.}
\vspace{-2em}
\label{fig:pipeline}
\end{figure*}

We propose \ours (\modelfull), a novel approach for fast and generalizable manipulation strategy learning from a single demonstration. \fig{fig:pipeline} shows the overall framework. First, we extract the contact points among objects from the reference trajectory. Subsequently, we find the candidate contact points that support the functional affordances on unseen test objects through a global-to-local matching process, in which we first perform coarse correspondence point matching using pretrained vision transformer (ViT) features DINOv2~\cite{oquab2023dinov2} (Section~\ref{subsec:dino}), followed by a local alignment based on shape curvatures to find stable local contact patches (Section~\ref{subsec:curvature}). Finally, the candidate contact points and their matching scores can be used for tool selection, for retargeting reference object trajectories, or as waypoints for motion planning (Section~\ref{subsec:transfer}). The generated trajectories will be verified in a physical simulation and then output to the robot for execution.

\subsection{Problem Definition}
\label{subsec:problem_definition}

We introduce the task of one-shot manipulation strategy learning. Specifically, we start with a single demonstration involving the $\text{SE}(3)$ trajectories of objects that execute a particular manipulation strategy to achieve a goal (\eg, hanging a hanger) in the reference scene. Our goal is to generate the trajectories of objects in a target scene that apply a similar manipulation strategy to a different object (\eg, attaching a mug to a mug tree). We simplify this problem by assuming the robot is only manipulating one of the objects (\ie, the ``tool'' that the robot is holding), and the desired motion can be explained by achieving a sequence of contact waypoints. Many rigid-body tool-using tasks such as hooking, poking, stacking, and funneling, all fall into this category.
For now, we assume that the strategy to apply and the manipulanda in both the reference and target scenes (\eg, the hanger or the mug) are already identified, and later we will discuss strategies for selecting the best tool object to accomplish a certain task.
Therefore, the problem can be cast as making the analogy between the reference object motion and the target object motion, taking into consideration other environmental constraints such as robot reachability and collisions.

Based on our insights into decomposing manipulation trajectories into contact point sequences, this task further reduces to establishing a functional correspondence between the demonstrated object and the novel object, which can be represented as an $\text{SE}(3)$ transformation between two rigid bodies for each contact waypoint. For tasks considered in this paper, we only consider scenarios where there is a single contact waypoint responsible for the target motion (but it generalizes to finding correspondences in multiple waypoints), and there is a reduced degree of freedom that can be reasoned about in 2D. In particular, we only model contacts on a canonical 2D cross-sectional view of the objects. For example, when we consider hooking objects on a table, we only consider object motions and shape properties on the surface that is perpendicular to the table's surface normal. This is equivalent to assuming access to a canonical pose of the object: for hooks, this will be the top-down view when the hook is placed on the table; for mugs, this will be the side view that contains the handle of the mug. In this paper, we assume that this canonical view is given, and in general, it can be predicted by external modules. Therefore, now, our goal is to establish a single correspondence between two 2D images of the objects in $\text{E}(2)$ (translations, rotations, and reflections in 2D) plus scaling; and then we can recover the $\text{SE}(3)$ correspondences based on the canonical object poses.

Overall, the input to \model is an image $I_T$ of the reference object $T$, an image $I_{T'}$ of the target object, and a point of interest $p_T$ on the reference image. Our goal is to find the corresponding point $p_{T'}$ on $I_{T'}$. When we have two objects interacting: the tool object $T$ being directly held by the robot (\eg, the hook) and another object $O$ in contact with $T$ (\eg, the target object we want to hook), our goal would be to find a contact point pair $p_{T'}$ on $I_{T'}$ and $p_{O'}$ on $I_{O'}$ given the reference $\left(I_T, I_O, p_T, p_O \right)$ that maximizes a score function: $\textit{score}\left( X, X' \right)$, where $X=\left( I_T, I_O,p_T,p_O\right)$ is the reference contact and $X'=\left( I_{T'}, I_{O'},p_{T'},p_{O'}\right)$ is the target contact. In this paper, we employ a two-stage global-to-local matching process, therefore, the $\textit{score}$ function will be composed of two parts: a global matching score $s_\textit{dino}$, and a local matching score $s_\textit{curv}$:
\[\textit{score}\left( X, X' \right) = s_\textit{dino}(I_T, I_{T'},p_T,p_{T'}) + \lambda \cdot s_\textit{curv}(X, X'),\]
where $\lambda$ is a constant hyperparameter. $p_{O}$ and $p_T$ are automatically extracted using the contact information in simulation and will be manually annotated for real-world objects (they can also be automatically recovered using video-based contact point detector such as \cite{ehsani2020use}). After generating top candidates for $p_{T'}$ and $p_{O'}$ based on $\textit{score}$, for additional validation, we will simulate the computed trajectory with the target object and select the first one that succeeded in the task in simulation. All tasks studied in this paper involve only stable quasi-static motion. Therefore, they can be simulated using mild assumptions on point clouds, uniform object densities, and frictions. When the simulation is unavailable in challenging dynamic tasks (\eg, hammering), we fall back to using the highest-scoring contact match.

\subsection{Global Contact Point Matching with Pretrained Features}
\label{subsec:dino}

\begin{figure*}[tp]
\centering
\includegraphics[width=\linewidth]{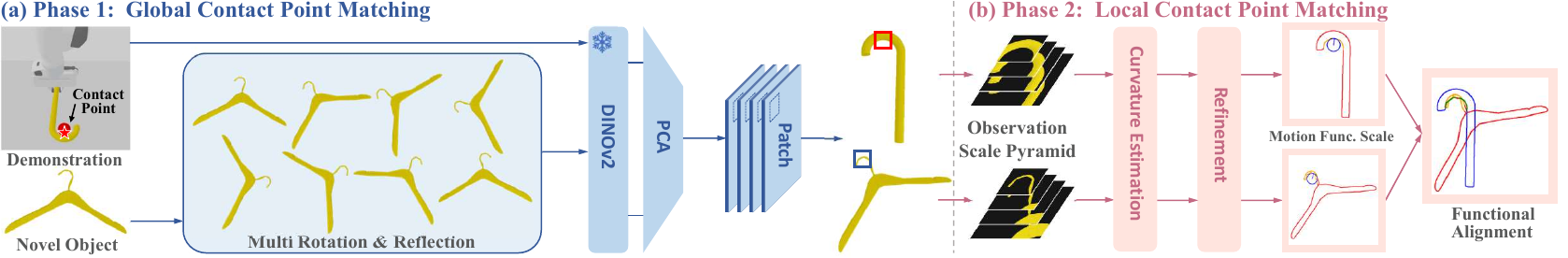}
\vspace{-1em}
\caption{\textbf{Global and Local Contact Point Matching.} The contact point matching process consists of two phases: (a) global matching using DINOv2~\cite{oquab2023dinov2} features; (b) local matching involving multi-scale curvature estimation and refinements using irrelevant point suppression and convexity matching.}
\label{fig:method}
\vspace{-2em}
\end{figure*}

The global matching score $s_\textit{dino}$ is only computed between the tool objects $T$ and $T'$. In this stage, we will find a candidate set of contact points ``globally'' on the target object $T'$, leveraging visual features pretrained on large image datasets for capturing global and semantic correspondences~\cite{zhang2023tale}. Specifically, we adopt DINOv2~\cite{oquab2023dinov2} as our visual feature extractor. \fig{fig:method}a shows the pipeline: we first extract visual features for both objects (with different image rotations and reflections) and then find a candidate set of matching points.

\xhdr{Feature extraction.} We feed $I_T$ and $I_{T'}$ into DINOv2 respectively, and get feature maps $F_T$ and $F_{T'}$. In practice, to eliminate the effects of rotation and reflection in 2D, we apply horizontal flips and 12 rotations on $I_{T'}$ and get a total of 24 images $\{F_{T'}^i\}_{i=1}^{24}$. Next, we apply a principal component analysis (PCA) on the feature vectors. Given feature maps $F_T, F_{T'}^i \in \mathbb{R}^{n \times n \times d}$, let $\textit{flatten}(F)$ be flattened version of $F$ in $\mathbb{R}^{n^2 \times d}$, we have
$W = \text{PCA}_{d,d'}\big \{ \textit{flatten}(F_T), \{\textit{flatten}(F_{T'}^i)\}_{i=1}^{24}\big \},$
where $n$ is the spatial resolution, $d$ and $d'$ are the original feature dimension and the feature dimension after reduction, respectively, and $W \in \mathbb{R}^{d\times d'}$ is the matrix of principal component vectors. The output is $\widetilde{F} = F \times W\in \mathbb{R}^{n\times n\times d'}$.

\xhdr{Matching by patches.} Next, we aggregate the features within a local region (instead of a single point) so as to increase the receptive field.
Formally, let $\textit{patch}( p )$ denote the $m \times m$ local area centered at point $p$, we find a point across all rotations of the target image $p_{T'}$ that maximizes the patch-aggregated cosine similarity with the reference point $p_T$:
\[
    s_\textit{dino}\left( I_T, I_{T'}, p_T, p_{T'} \right) = \sum_{\genfrac{}{}{0pt}{1}{p \in \textit{patch}(p_{T}),}{p' \in \textit{patch}(p_{T'})}} \left\langle\widetilde{F}_T(p), \widetilde{F}_{T'}^{i}(p') \right\rangle.
\]
We select the top $k=3$ matches with the highest $s_\textit{dino}$ to perform the local contact point matching in the next stage.

\subsection{Local Contact Point Matching with Curvature Estimation}
\label{subsec:curvature}

Although DINOv2 can propose contact points within a coarse global region, it is not accurate enough in terms of local geometric properties to support collision-free and stable physical motion.
We leverage curvatures to refine the correspondence matching.  
Illustrated in \fig{fig:method}b, our local alignment has four steps. We first construct a multiple observation scale pyramid: we estimate the curvatures of the reference contact point and the contact point proposed by DINOv2 on the target object at multiple scales. Then, based on the normal direction and the sign of the estimated curvatures, we perform irrelevant point suppression and convexity matching to find a physically plausible contact point for each scale. After that, we repeat the curvature estimation process on the updated contact points to get a more accurate curvature and normal direction. Finally, we accept the best contact point across scales.

\xhdr{Curvature estimation at a given scale.} Given an image of the object with segmentation, we first use the Canny edge detector~\cite{canny1986computational} to get the object edges. We define the \textit{observation scale} as the radius of the region centered at the point of interest (e.g., the contact point) on the object edge. For a certain observation scale $s$ of the point $\mathbf{c}$, we can use the edge points $\{(x_i, y_i)\}_{i=1}^{n}$ inside $s$ to estimate the magnitude $\kappa$ of the curvature and the radius of curvature $r$. Specifically, we first find the direction $\mathbf{x'}$ with the largest variance on $\{(x_i, y_i)\}_{i=1}^{n}$, and construct a local coordinate system $\mathbf{x'y'}$ centered at $\mathbf{c}$, then represent $\{(x_i, y_i)\}_{i=1}^{n}$ in $\mathbf{x'y'}$ to get $\{(x_i', y_i')\}_{i=1}^{n}$. Afterwards, we fit a parabola $y' = ax'^2$ on points $\{(x_i', y_i')\}_{i=1}^{n}$, i.e.,
\[
    a = \arg \min_a \sum_{i=1}^{n} \left(y_i' - ax_i'^2\right)^2,
\]
then by the definition of curvature, we have 
$\kappa = 2|a|, r = 1/\kappa. \label{eq:curvature}.$

Now, for a pyramid of observation scales $\{s_j\}_{j=1}^{m}$, we can compute a series of radii of curvature $\{r_j\}_{j=1}^{m}$. The \textit{motion functional scale} is defined as 
$s_j, \text{where}~j = \arg \min_{j'} \left|\frac{s_{j'}}{r_{j'}} - \alpha \right|$,
where $\alpha$ is a parameter, and in practice, we select $\alpha = 3.5$ for all objects across all experiments. This constant corresponds to a scenario where the observation scale is similar to the radius of curvature estimated using the points inside the observation scale. Furthermore, the sign of the curvature (that is, whether the point on the curve is convex or concave) can be computed based on the mask of the object. Given both the tool object $T'$ and target object $O'$, we define the local matching score as:
\[s_\textit{curv}\left(\langle I_T,I_O,p_T,p_O \rangle, \langle I_{T'},I_{O'},p_{T'},p_{O'} \rangle\right) = \left| \frac{r(p_T)}{r(p_O)} - \frac{r(p_{T'})}{r(p_{O'})}\right|. \]

\begin{figure}[tp]
\centering
\includegraphics[width=\linewidth]{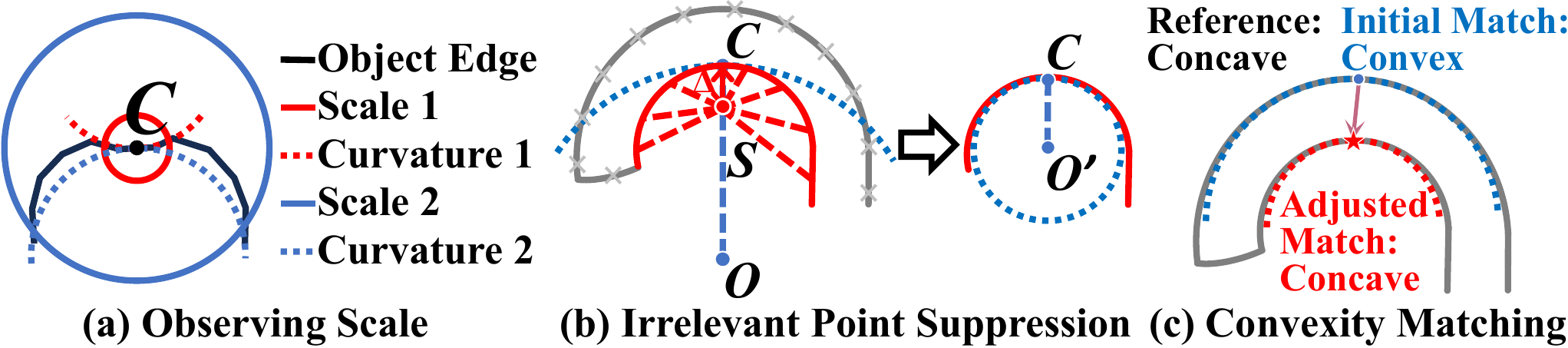}
\vspace{-1em}
\caption{\label{fig:local_match}
\textbf{(a)} The observing scale at which we perform estimation has effect on both the magnitude and the sign of the estimated curvature. \textbf{(b)} Filtering out the irrelevant points can give a more accurate curvature estimation. \textbf{(c)} We perform local search within the observation region to find the contact point with the correct convexity. 
}
\vspace{-2em}
\end{figure}

\xhdr{Irrelevant point suppression and convexity matching.}
Direct application of the curvature estimation algorithm is not robust enough for two reasons. First, the choice of the observation scale will significantly affect the estimation of the curvature. As illustrated in \fig{fig:local_match}a, selecting a very small scale (scale 1) will make the estimation sensitive to local noise. Second, for large observation scales, irrelevant points (\ie, points on a different ``edge'' of the object) will be included and inject noise in the estimation, especially for thin objects as shown in \fig{fig:local_match}b.

To eliminate these irrelevant points around the contact point of interest, we compute the curvature twice. First, we include all points within the observation scale. After computing the initial contact point $C$ on the edge of the object and its curvature, we pick a point $S$ that is at a small distance $\Delta$ from the contact point $C$ along the direction of the radius of curvature $\overrightarrow{CO}$ in the initial estimation. Then, we select all points on the same ``edge'' as $C$ on the side of $S$. This is done by emitting ``rays'' in all directions from $S$ and for each ray selecting the point that is hit first. We then use these points to estimate the curvature again, as shown on the right of Figure~\ref{fig:local_match}b. Moreover, we ensure that the curvature sign (\ie, the convexity) of the corresponding point in the target image is consistent with that in the reference image by performing a local search within the observation region of the target image, which is depicted in \fig{fig:local_match}c.

\subsection{Generating Object Motions by Making Contact Analogies}
\label{subsec:transfer}
So far we have presented a general mechanism for finding geometric functional correspondence between reference and target objects, and it can be used in many downstream modules. In the one-shot manipulation strategy learning setting, we consider three particular use cases.

\xhdr{Object motion retargeting.} This is the most straightforward way to generate target object motions assuming there is a single contact that accounts for the motion. Given the reference contact point $p_T$, the direction vector $\mathbf{v}$ of the view angle, the normal vector $\mathbf{n}_T$ of $p_T$ estimated from curvature calculation, we can construct two local frames for the reference and the target object, and directly retarget the reference object motion by aligning two local frames.

\xhdr{Motion planning based on contact waypoints and force directions.} We can generate object motion by leveraging analogical contact points as waypoints for a motion planner. For example, if our goal is to hang an object at a particular position, we can first compute the hanging pose by making analogies with the reference object pose and use a collision-free motion planner to generate the robot motion. This also applies to hook-using tasks where the goal is specified as to apply a particular force along a target direction on an object.

\xhdr{Tool selection.} Our pipeline also supports tool selection, where the goal is to select one ``tool'' object (\eg, a hook) from a set of available objects that can best execute the strategy in the new scenario, given the other object to manipulate (\eg, the object to be hooked). We apply our algorithm on all available tools, and select the tool with the highest score.

\section{Experiments}\label{sec:experiments}
In this work, we have conducted experiments both in simulation and in the real world. For simulation experiments, we adopt the simulation environment SAPIEN 2~\cite{xiang2020sapien} using a Franka Emika Panda arm. We evaluate different one-shot manipulation strategy learning algorithms on three tasks, as shown in \fig{fig:diverse_objects}: \textbf{(\textit{Scooping})} scooping balls of different sizes with various spoons against a concave arc, given a demonstration with a reference spoon; \textbf{(\textit{Hanging})} hanging a mug onto a mug tree, given a demonstration of hanging a hanger on a rod; \textbf{(\textit{Hooking})} selecting a tool from a set to hook objects of varying shapes and sizes, given a demonstration of hooking a ball with a hook. To generate object motions, we adopt object motion retargeting for \textit{Scooping} and motion planning for \textit{Hanging} and \textit{Hooking} from Section~\ref{subsec:transfer}. For all tasks, we have two variants: {\it Floating-Gripper (FG)} where the tool will be manipulated by a floating gripper, and the harder {\it Arm} variant where the tool needs to be grasped and manipulated by the robot. In this variant, we use a general antipodal grasp sampler~\cite{nguyen1985synthesis} and RRT-Connect~\cite{lavalle2001randomized} for generating robot trajectories, of which we utilize MPlib~\cite{su2023mplib} as the implementation. For \textit{Hooking}, we also have a variant where we provide a random tool to the floating gripper (no tool selection). For all tasks, we only provide a single demonstration of object motion trajectories, the canonical view, and the pair of contact points on reference objects.

\xhdr{Baselines.} We implemented two sets of baselines: global shape-matching and local shape-matching methods. For global shape-matching methods, we use two different methods: principal component analysis (PCA) and iterative closest point with FPFH features (FPFH+ICP)~\cite{rusu2009fast, arun1987least, zhou2018open3d} on object point clouds to find the best transformation that would align the target object and the reference object. Then, we retarget the object motion or transform the waypoints.
For local shape-matching baselines, we also implemented two methods: {\it DINO matching} which directly finds the best functional correspondence based on the DINOv2 features, and {\it curvature matching} which uses the random contact point sampling method from \cite{mao2023learning} and applies the curvature filtering algorithm from \model to rank all contact points. Since this method involves random sampling of contact points and simulation, we cap its runtime to 180 seconds and use the contact point with the highest matching score found at that time. Note that all methods except for the sampling-based curvature-matching algorithm have very small variances across different runs (our algorithm is almost deterministic). Therefore, we do not include performance variances for different methods. We do not compare with methods such as NDF~\cite{simeonov2022neural}, KETO~\cite{qin2020keto}, and GIFT~\cite{turpin2021gift} in the main experiment, because they assume access to a fairly large dataset of 3D object models for representation learning, but we provide a case study of comparing \ours with KETO in Table~\ref{tab:5}.

\subsection{Experiment Results in Simulation}

\begin{table}[tp]
\centering\small
    \begin{adjustbox}{width=\linewidth}
    \setlength{\tabcolsep}{3.5pt}
\begin{tabular}{lccccccc}
    \toprule
    \multirow{2}{*}{\textbf{Method}} & \multicolumn{2}{c}{\textbf{Scooping}} & \multicolumn{2}{c}{\textbf{Hanging}} & \multicolumn{3}{c}{\textbf{Hooking}} \\
    \cmidrule(lr){2-3} \cmidrule(lr){4-5} \cmidrule{6-8}
     & FG & Arm & FG & Arm & No Tool Sel. & FG & Arm\\
    \midrule 
    \textbf{PCA} & 27.6 & 16.7 & 0.0 & 0.0 & 5.0 & 0.0 & 0.0 \\
    \textbf{FPFH+ICP} & 32.9 & 27.2 & 0.0 & 0.0 & 6.7 & 0.0 & 0.0 \\
    \textbf{DINO Matching} & 55.7 & 41.4 & 23.9 & 21.6 & 49.2 & 66.7 & 24.0 \\
    \textbf{Curvature-Only} & 13.9 & 5.6 & 68.7 & 38.1 & 60.0 & 76.7 & 3.3 \\
    \textbf{\ours} & \textbf{94.4} & \textbf{65.4} & \textbf{92.9} & \textbf{84.3} & \textbf{65.0} & \textbf{100} & \textbf{63.3} \\
    \bottomrule
\end{tabular}

    \end{adjustbox}
    \vspace{0.05em}
    \caption{\textbf{Average success rates (\%) of 3 tasks.} `\textit{floating gripper} (FG)' indicates manipulation by directly setting positions and velocities of objects, focusing only on object-object interactions, excluding the robot. All methods except for Curvature-Only run almost deterministically. Curvature-Only has an average std. of 6.6\%.}
    \label{tab:1}
    \vspace{-2.5em}
\end{table}

\tbl{tab:1} summarizes the overall performances of variants of our methods and other baselines. The success rates are computed over 6 spoons and 3 objects for \textit{Scooping}, 134 mugs for \textit{Hanging} adopted from \cite{simeonov2022neural}, and 4 tools and 5 objects for \textit{Hooking} adopted from \cite{mao2023learning}. Overall, our model \model performs the best. We break down our analysis into the following bullet points.

\begin{figure*}[tp]
\centering
\includegraphics[width=\linewidth]{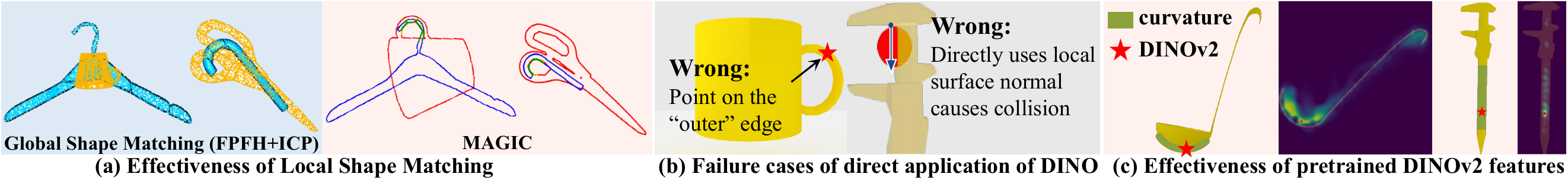}
\vspace{-1em}
\caption{\textbf{(a)} Global shape matching fails to provide effective functional alignment due to the lack of awareness of local contact points. \textbf{(b)} Local curvature helps identify more accurate contact points and contact normals. \textbf{(c)} Pretrained DINOv2 features enhance data efficiency by selecting the optimal contact (the red star) and grasping points from those that curvature alone cannot distinguish (the green region). We also visualize the DINOv2 matching heatmaps.}
\label{fig:exp1}
\vspace{-1.5em}
\end{figure*}

\xhdr{Local shape matching can significantly improve performance over global shape matching.} We compare \ours with global shape matching methods, PCA and FPFH+ICP. As shown in Table~\ref{tab:1}, global shape matching is ineffective because it lacks an understanding of the local contact points. \fig{fig:exp1}(a) illustrates an example of the correspondence established by global shape matching (left) and \ours (right). To further demonstrate the effectiveness of local shape matching, we also provide an additional study on the mug-hanging task in Table~\ref{tab:2}, where the demonstration includes the trajectory of hanging a held-out mug on the mug tree (\ie, intra-category transfer), rather than the hanger. In this setting, FPFH+ICP achieves a success rate of 41.1\% (\cf 94.0\% for \model). 
The results indicate that PCA and FPFH+ICP (as well as the DINO matching algorithm) can sometimes successfully achieve intra-category transfer by aligning overall shapes. However, they perform poorly at inter-category transfer due to the lack of local contact point alignment. By contrast, \ours excels in both intra-category and inter-category transfers by accurately identifying contact points that satisfy both global and local constraints.

\begin{table}[htbp]
    \centering
    \vspace{-0.05em}
    \begin{adjustbox}{width=\linewidth}
    \begin{tabular}{lccccc}
\toprule
& \textbf{PCA} & \textbf{FPFH+ICP} & \textbf{DINO Match} & \textbf{Curvature-Only} & \textbf{\model} \\
\midrule
Hanging (intra-cat.) & 5.2 & 41.1 & 64.9 & 67.9 & \textbf{94.0} \\
Hanging (inter-cat.) & 0.0 & 0.0 & 23.9 & 68.7 & \textbf{92.9} \\
\bottomrule
\end{tabular}

    \end{adjustbox}
    \vspace{0.05in}
    \caption{\textbf{Average success rates (\%) of \textit{Hanging} (intra-category transfer vs. inter-category transfer).} While PCA and FPFH+ICP can occasionally achieve intra-category transfer, they fail to perform inter-category transfer. The experiments are conducted on the \textit{FG} variant.}
    \label{tab:2}
    \vspace{-1em}
\end{table}

\xhdr{Local curvature matching is crucial, in addition to pretrained features.} The comparison between \model and other methods that do not use local curvature information (\eg, PCA, FPFH+ICP, and DINO Match) shows the importance of local curvature matching. As an illustration, as shown in \fig{fig:exp1}(b), the contact points proposed solely by the pretrained model are likely to fail due to incorrect curvature sign and incorrect interaction direction. By contrast, \ours leverages local curvature information to identify the appropriate contact points and correct force direction through the curvature estimation.  Moreover, on the \textit{Hooking} task, pairwise curvature matching plays an important role in selecting the most appropriate tool for different objects, as shown in the comparison between the {\it No Tool Sel.} and the {\it FG} variant in the hooking task. 

\xhdr{Using pretrained features as guidance significantly improves contact point selection.} In Table~\ref{tab:1}, \ours shows superior performance over the sampling-based curvature matching algorithm. In principle, without noise in curvature computation and given a sufficient number of sampled points, the curvature-based method would be capable of finding the best contact. However, it will be very slow (\eg, in \cite{mao2023learning}, the authors set the timeout to 600 seconds for the search). In this work, we leverage the global and semantic correspondence provided by DINOv2 features to reduce the searching space of local geometry matching.
In \fig{fig:exp1}(c), we visualize how pretrained DINOv2 features can improve data efficiency by identifying the best contact and grasping points that curvature alone cannot differentiate.

Moreover, we provide additional comparison of different pre-trained features such as DIFT~\cite{tang2023emergent} and SD-DINO~\cite{zhang2023tale} in Table~\ref{tab:4}. The results indicate that DINOv2 not only provides better guidance for shape matching but it also runs faster. This difference can be attributed to the distinct pretraining tasks and model architectures of DINOv2. DINOv2 employs self-supervised learning techniques, involving contrastive learning and a teacher-student training paradigm, where the teacher network has access to global views and distills knowledge to the student network --- which has access to both global and local views~\cite{oquab2023dinov2}. This approach enhances the model's understanding of global information. By contrast, diffusion models are trained on generation tasks through a denoising process, which focuses more on local textures~\cite{ho2020denoising}. Regarding model structures, DINOv2 uses Vision Transformer (ViT)~\cite{dosovitskiy2021image} with an attention mechanism applied to patches of the input image, whereas diffusion models are based on U-Net~\cite{ronneberger2015unet} and process the input and output as full-resolution images, requiring more computation time.

\begin{table}[htbp]
    \centering\small
    \begin{adjustbox}{width=\linewidth}
    \begin{tabular}{lcccc}
\toprule
\textbf{Method} & \textbf{Scooping} & \textbf{Hanging } & \textbf{Hooking} & \textbf{Avg. Time}\\
\midrule
\ours (DIFT + Curvature) & 62.2 & 81.3 & \textbf{100} & 60.8s \\
\ours (SD-DINO + Curvature) & 73.8 & 91.8 & 96.0 & 50.7s \\
\ours (DINOv2 + Curvature) & \textbf{94.4} & \textbf{92.9} & \textbf{100}  & \textbf{34.0s}\\
\bottomrule
\end{tabular}

    \end{adjustbox}
    \vspace{0.05in}
    \caption{\textbf{Additional studies on the choice of pretrained image features.} We substitute DINOv2 with SD-DINO~\cite{zhang2023tale} and DIFT~\cite{dosovitskiy2021image}. We present the average success rates (\%) and average time consumed of the methods. The experiments are conducted on the \textit{FG} variant.}
    \label{tab:4}
    \vspace{-1em}
\end{table}

\xhdr{Further comparison with dataset-dependent methods.} KETO~\cite{qin2020keto} and GIFT~\cite{turpin2021gift} introduce keypoints as the guidance for tool-using strategies and utilize self-supervised learning on the positive samples obtained from trial-and-error with the tools from a large tool dataset. While \ours also introduces a similar concept, contact points, we propose to consider contacts as pairs of contact points. In contrast, keypoints only focus on the tool object. Moreover, both KETO and GIFT need a dataset of tools, while \ours only requires one single demonstration and can be generalized to tools with very different shapes.

In Table~\ref{tab:5}, we provide additional experiment results comparing \ours with KETO over the \textit{Hammering} and \textit{Pushing} tasks from KETO. Since KETO requires a decently large dataset of tool meshes to perform self-supervised learning, which is not available for those tasks in our main experiment, we perform experiments on their tasks. The single demonstration for \ours is composed of an image of a reference hammer taken in the simulator and the pixel coordinates of the function point and the effect point. The results indicate that \ours outperforms KETO on both tasks. To further investigate the generalizability, we provide the success rate of transferring from hammers to hammers and non-hammers, respectively. For both intra-category and inter-category generalization, \ours performs better than KETO.

\begin{table}
    \centering\small
    \begin{adjustbox}{width=\linewidth}
    \setlength{\tabcolsep}{13.5pt}
\begin{tabular}{lcccc}
    \toprule
    \multirow{4}{*}{\textbf{Method}} & \multicolumn{3}{c}{\textbf{Hammering}} & \textbf{Pushing}\\
    \cmidrule(lr){2-4} \cmidrule(lr){5-5}
    & \makecell{Hammers $\rightarrow$ \\ Hammers} & \makecell{Hammers $\rightarrow$ \\ Non-hammers} & Average & Average \\
    \midrule 
    \textbf{KETO} & 66.4 & 46.3 & 56.4 & 71.7 \\
    \textbf{MAGIC} & \textbf{75.6} & \textbf{53.7} & \textbf{64.4} & \textbf{76.3} \\
    \bottomrule
\end{tabular}

    \end{adjustbox}
    \vspace{0.05in}
    \caption{\textbf{Comparison between MAGIC and KETO~\cite{qin2020keto} (Learning Keypoint Representations for Tool Manipulation).}  We present the average success rates (\%). We also investigate the intra-category and inter-category generalizablity of \ours and KETO, indicated by Hammers $\to$ Hammers and Hammers $\to$ Non-hammers, respectively.}
    \vspace{-1em}
    \label{tab:5}
\end{table}

\label{subsec:ablation_studies}
\begin{table}[tp]
\vspace{-0.05em}
    \centering
    \begin{adjustbox}{width=\linewidth}
    \begin{tabular}{lcccc}
\toprule
\textbf{Method} & \textbf{Scooping} & \textbf{Hanging} & \textbf{Hooking} & \textbf{Avg. Time}\\
\midrule
DINO Matching (Best Baseline) & 55.7 & 23.9 & 66.7 & \textbf{12.4s} \\ 
\ours (DINOv2 + Curvature) & \textbf{94.4} & 92.9 & \textbf{100}  & 34.0s\\ \midrule
(w/o. Feature PCA and Patch) & 71.7 & \textbf{93.3}  & 62.0  & 33.8s\\
(w/o. Two-Step Curvature) & 83.3 & 53.7  & 75.6 & 14.1s\\
(w/o. Simulation Verification) & 77.8 & 71.6 & 50.0 & 19.7s \\
\bottomrule
\end{tabular}

    \end{adjustbox}
    \vspace{4pt}
    \caption{\textbf{Ablation studies.}  We present the average success rates (\%) and average runtime of all methods, on the \textit{FG} variant.}
    \label{tab:3}
    \vspace{-1em}
\end{table}

\xhdr{Ablation studies.} Table~\ref{tab:3} provides additional ablation studies on different design choices of our method \model. First, there is a significant performance drop observed in the scooping and hooking tasks if we remove feature PCA and patch-based score aggregation. This demonstrates that both methods reduce noise in the feature maps and improve the accuracy of contact point matching. Second, we consider removing the irrelevant point suppression and convexity matching. The model still surpasses vanilla DINO matching, which demonstrations the effectiveness of curvature estimation, but we see a clear performance drop compared with the full algorithm \model. Finally, we demonstrate that removing additional geometric and physical feasibility checks based on the simulator, and instead selecting only the trajectory with the highest matching score, results in a notable performance decline. While our matching pipeline and motion generator can generate plausible contact points and actions, simulation verification remains essential to ensure that the executed trajectory is collision-free and maintains physical stability.

\xhdr{}

\subsection{Real-World Experiments}

\begin{figure}[tp]
    \centering
    \includegraphics[width=\linewidth]{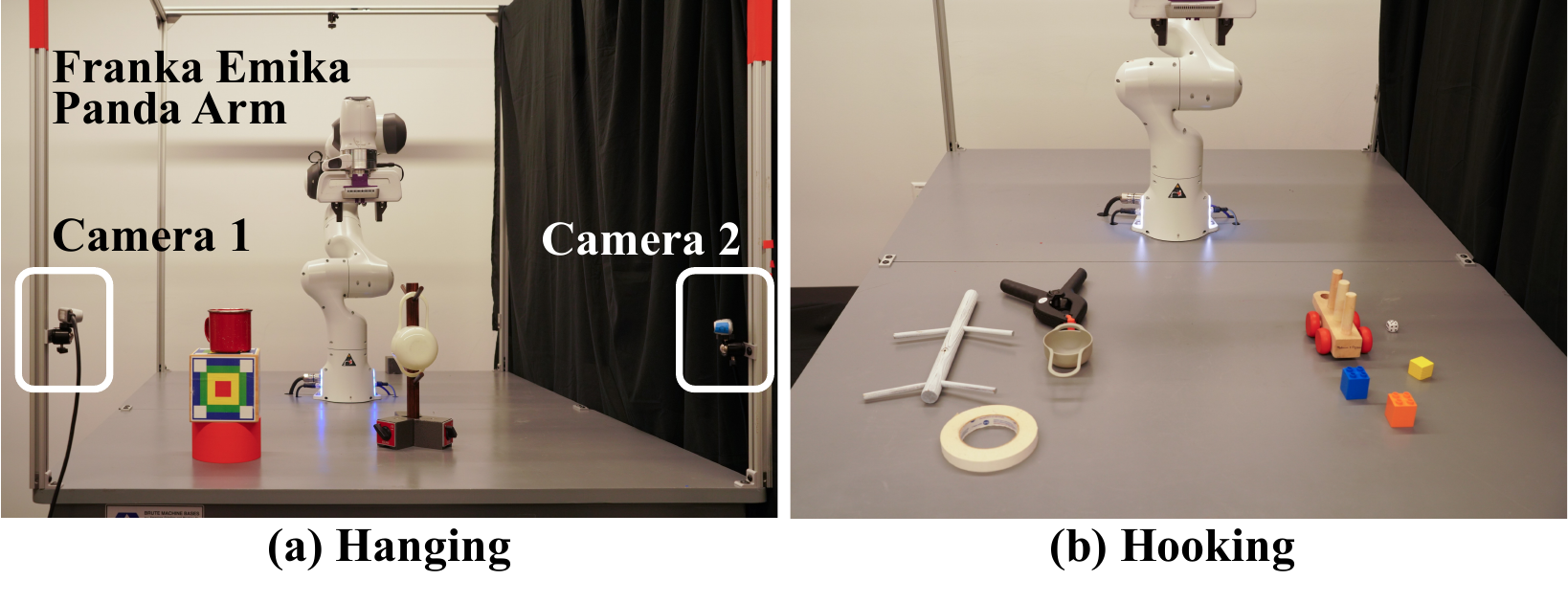}
    \vspace{-1em}
    \caption{\textbf{Real-world experiment setup.}}
    \label{fig:real_experiment_set_up}
    \vspace{-1em}
\end{figure}

\begin{figure}[tp]
    \centering
    \includegraphics[width=\linewidth]{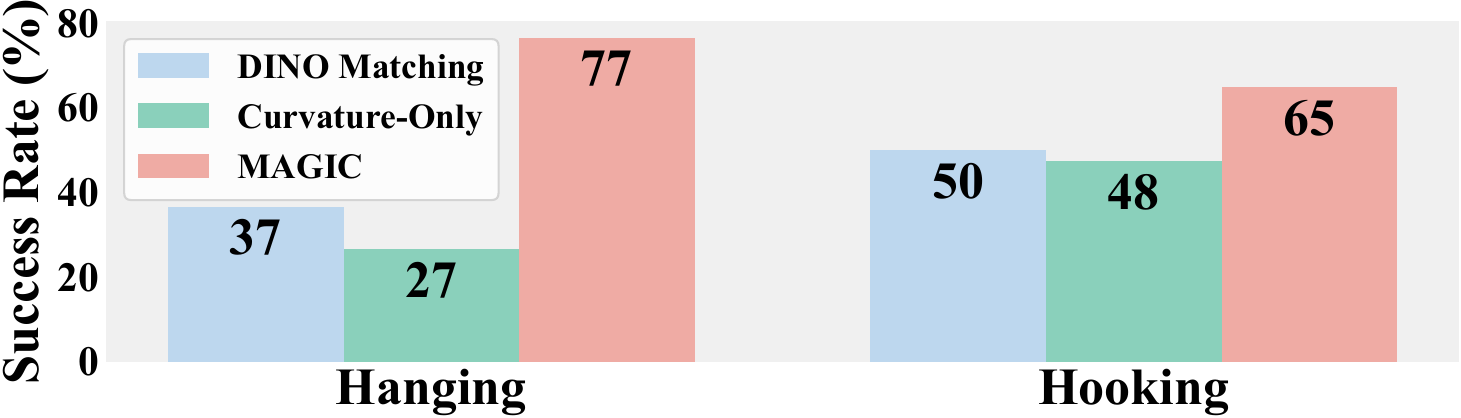}
    \caption{\textbf{Average success rates of real-world experiments.}}
    \label{fig:real_success_rate}
    \vspace{-1em}
\end{figure}

To evaluate the effectiveness of \ours in real-robot systems, we conducted experiments on \textit{Hanging} and \textit{Hooking} utilizing a Franka Emika Panda arm with a parallel gripper, as depicted in \fig{fig:real_experiment_set_up}. We capture the RGB and depth image of the scene with two RealSense D435 cameras, whose extrinsics are calibrated to the frame of the base link of the robot. 

We perform contact point matching on the RGB image captured by the camera, then import a reconstructed (possibly partial) object mesh from the RGBD camera into the SAPIEN 2 simulator for motion retargetting, motion planning, and verification. To reconstruct the mesh of the objects from the RGBD image, we first use the Segment Anything model~\cite{kirillov2023segment} to get the mask of the objects, and extract the corresponding point cloud of each object. Then, we remove outliers with DBSCAN~\cite{ester1996density} and select the cluster with the largest number of points. We then perform object completion by projecting the point cloud down to the table plane, and using Alpha Shape~\cite{edelsbrunner1983shape} to reconstruct the surfaces of the object mesh. 

\begin{figure}[tp]
\centering
\vspace{-0em}
\includegraphics[width=\linewidth]{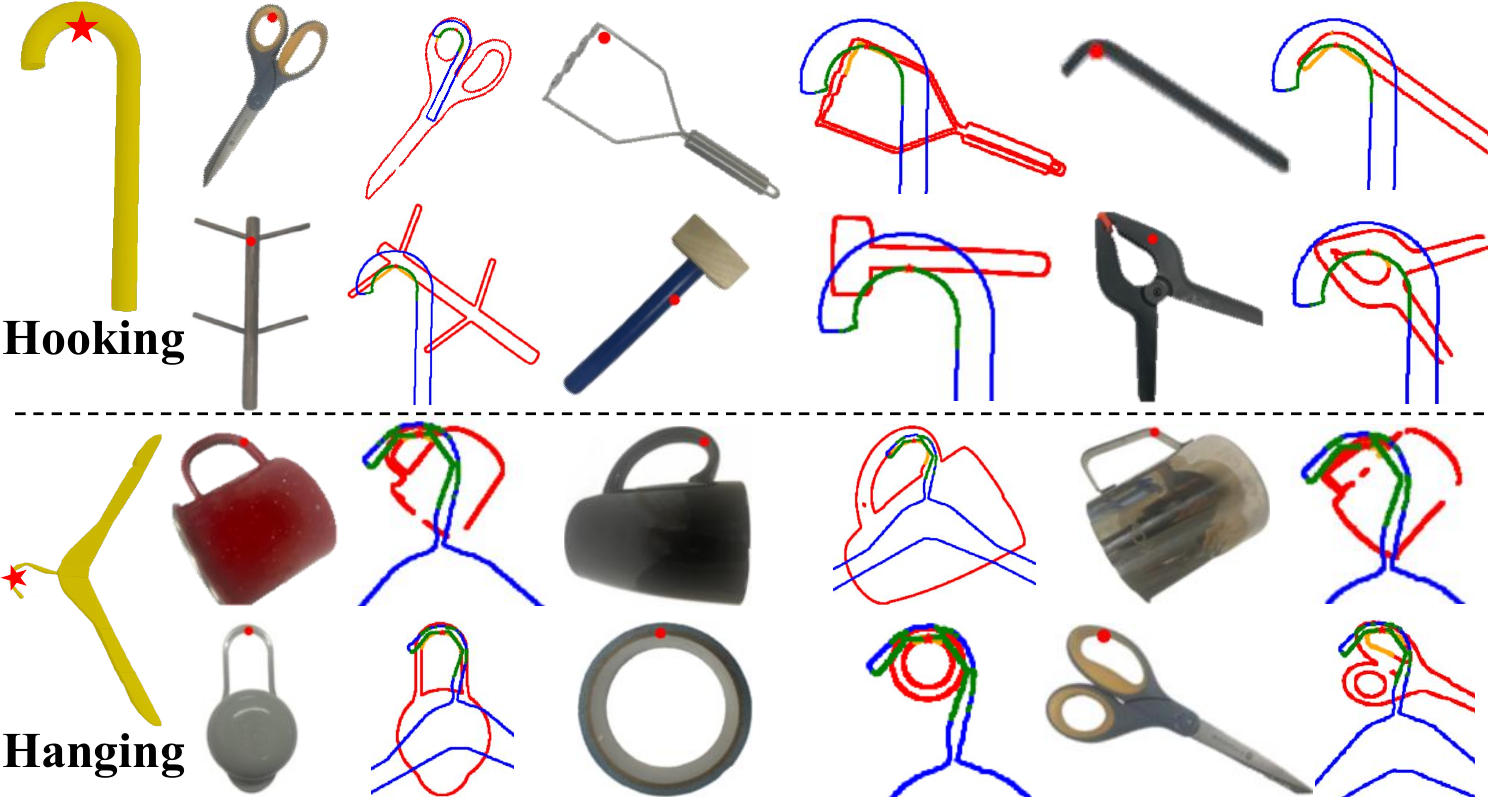}
\vspace{-1em}
\caption{\textbf{Contact point matching results on real-world objects.}}
\label{fig:real_demo}
\vspace{-1.8em}
\end{figure}

To provide a quantitative assessment, we perform a total of 10 trials for each object. We report the success rates of various methods for \textit{Hanging} across three visually distinct cups and \textit{Hooking} across four tools from different categories, as shown in \fig{fig:real_success_rate}. \ours has consistently achieved the highest performance, which validates the effectiveness and practical applicability of \ours in the real world. We also illustrate the contact point matching found by \ours on more real-world objects in \fig{fig:real_demo}. For failure case analysis, please check out our project \href{magic-2024.github.io}{website}.

\section{Conclusion}\label{sec:conclusions}
We have presented \ours, a framework for one-shot manipulation strategy learning, enabling robots to quickly adapt and execute tool-using tasks with novel objects. We integrate both data-driven and analytical shape-matching algorithms for the best of both worlds, to quickly generate precise and physically plausible contact points. Noteably, our algorithm is generic in the sense that we do not need task-specific information for the matching algorithm other than the reference contact waypoints. Experiments on three representative tasks illustrate the effectiveness of our method.

\xhdr{Limitations.} Currently, our pipeline is designed for making contact analogies for a single contact patch between two objects. Future work should consider interactions among multiple objects with multiple contact points, and extend the current shape-based matching criteria to consider forceful affordance. Another future direction would be to consider predicting or searching over 2D views of objects for correspondence matching. Moreover, our overall framework of coarse-to-fine and semantic-to-geometric matching can be extended to 3D pretrained features~\cite{shen2023distilled, wang2024d3fields, dutt2024diffusion} and curvatures. 

\xhdr{Acknowledgements.} We gratefully acknowledge support from NSF grant 2214177; from AFOSR grant FA9550-22-1-0249; from ONR MURI grant N00014-22-1-2740; and from ARO grant W911NF-23-1-0034; from MIT Quest for Intelligence; from the MIT-IBM Watson AI Lab; from the Boston Dynamics AI Institute; from ONR Science of AI; and from Simons Center for the Social Brain. Any opinions, findings, and conclusions or recommendations expressed in this material are those of the authors and do not necessarily reflect the views of our sponsors.

\begin{figure*}[hp]
    \centering
    \includegraphics[width=\textwidth]{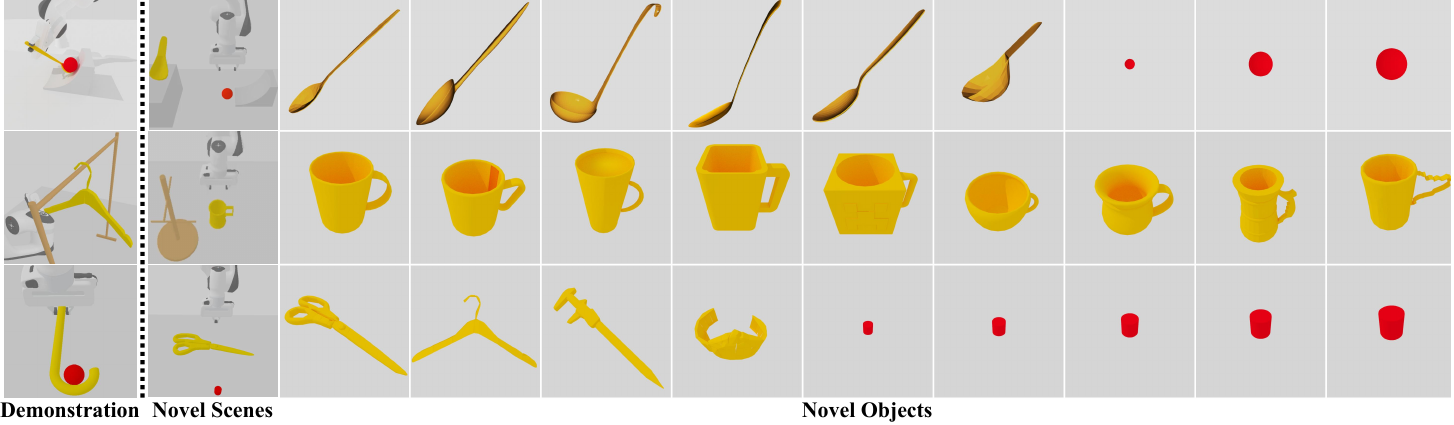}
    \caption{\textbf{Visualization of demonstrations and novel scenes/objects.} We evaluate the ability of one-shot manipulation strategy learning by providing a single demonstration for three tasks: \textit{Scooping}, \textit{Hanging}, and \textit{Hooking} (shown in the first column), and then testing on various unseen instances.}
    \label{fig:diverse_objects}
    \vspace{-6in}
\end{figure*}

\begin{figure*}
    \centering
    \begin{subfigure}[t]{0.495\linewidth}
        \centering
        \includegraphics[width=\linewidth]{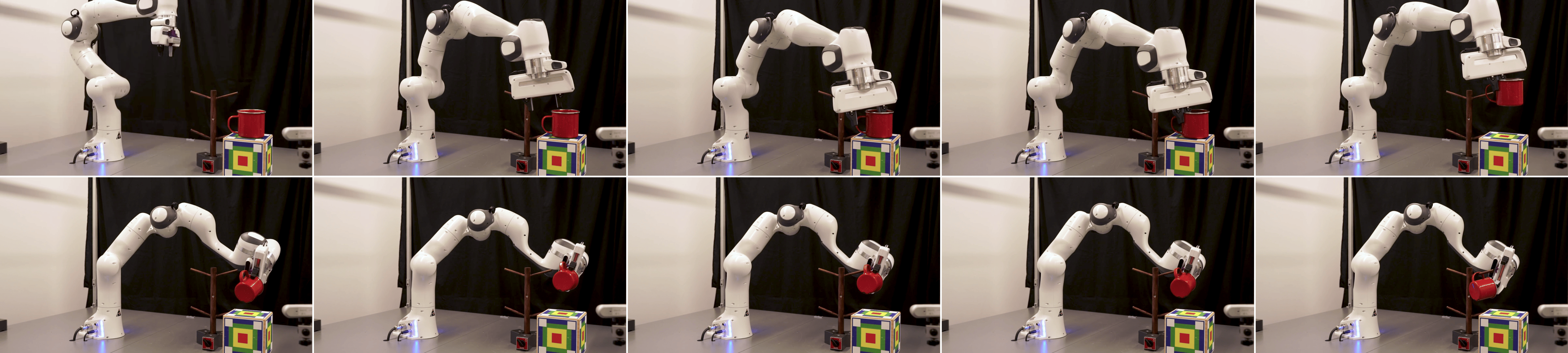}
    \end{subfigure}
    \hfill
    \begin{subfigure}[t]{0.495\linewidth}
        \centering
        \includegraphics[width=\linewidth]{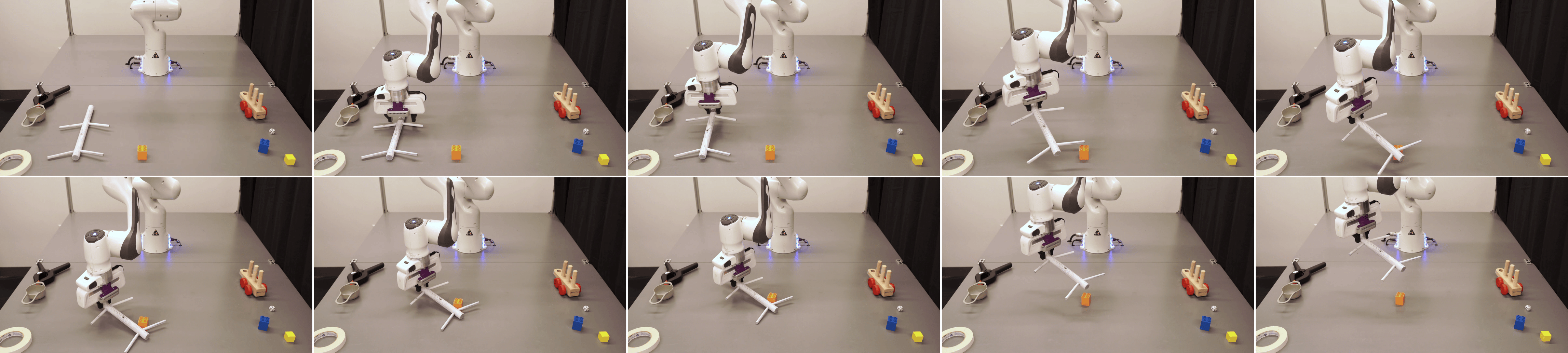}
    \end{subfigure}
    \caption{\textbf{Example episodes of \textit{Hanging} and \textit{Hooking} in the real world.}}
    \vspace{-1.5em}
    \label{fig:real_demo_combined}
\end{figure*}

\section*{Appendix}\label{sec:appendix}
\xhdr{Visualization of novel scenes and diverse objects}. In \fig{fig:diverse_objects}, we present the demonstrations, the novel scenes, and examples of novel objects for each task. \\
\textit{1) Scooping:} Demonstrated on a reference spoon and a reference ball; evaluated on 4 spoons, a soup ladle, a measuring cup, and 3 different balls. \\
\textit{2) Hanging:} Demonstrated on hanging a hanger to a rod; evaluated on hanging mugs to a mug tree~\cite{simeonov2022neural}. where the meshes of the mugs are adopted from ShapeNet~\cite{chang2015shapenet}. We filter out those meshes unsuitable for hanging (e.g., mugs without handles), and use 134 mugs for evaluation. \\
\textit{3) Hooking:} Demonstrated on hooking a reference object with a hook; Evaluated on hooking 5 different cylinders with 4 tools (a pair of scissors, a hanger, a caliper, and a watch).

\begin{figure}[htbp]
    \centering
    \includegraphics[width=\linewidth]{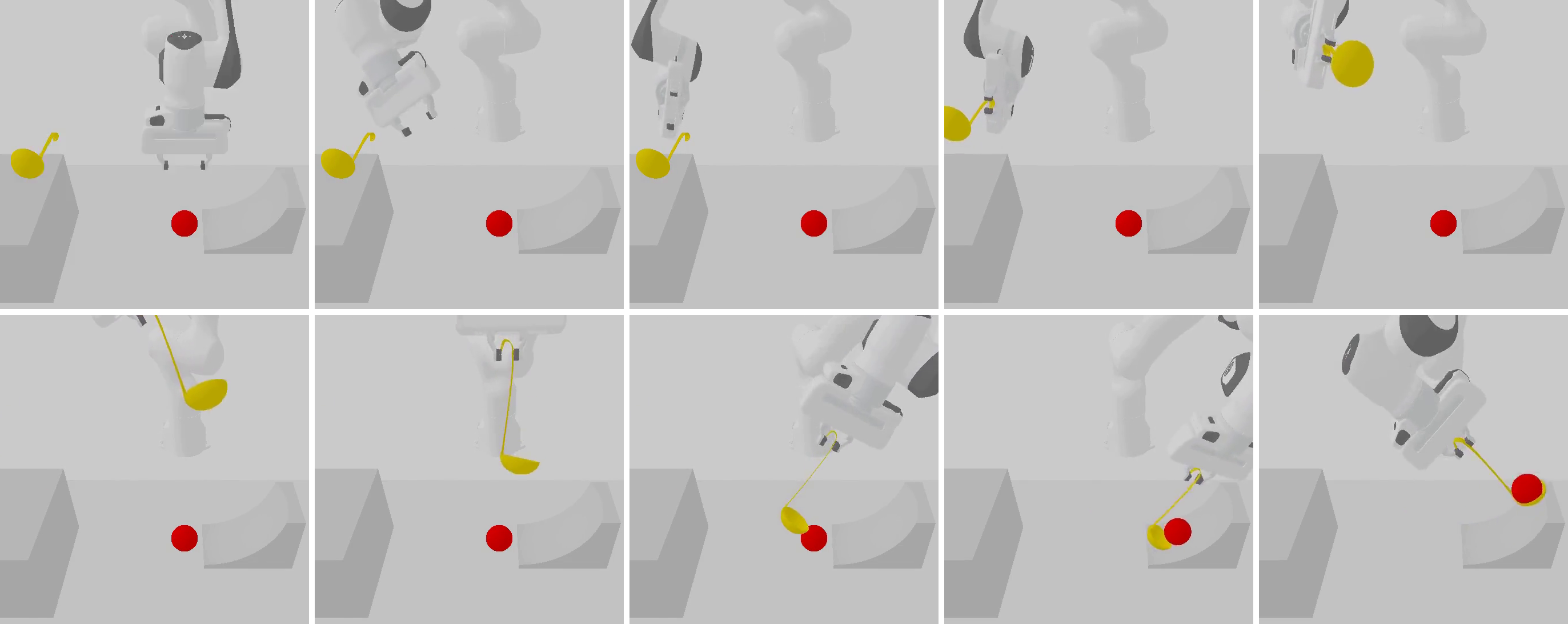}\\
    \vspace{0.05in}
    \includegraphics[width=\linewidth]{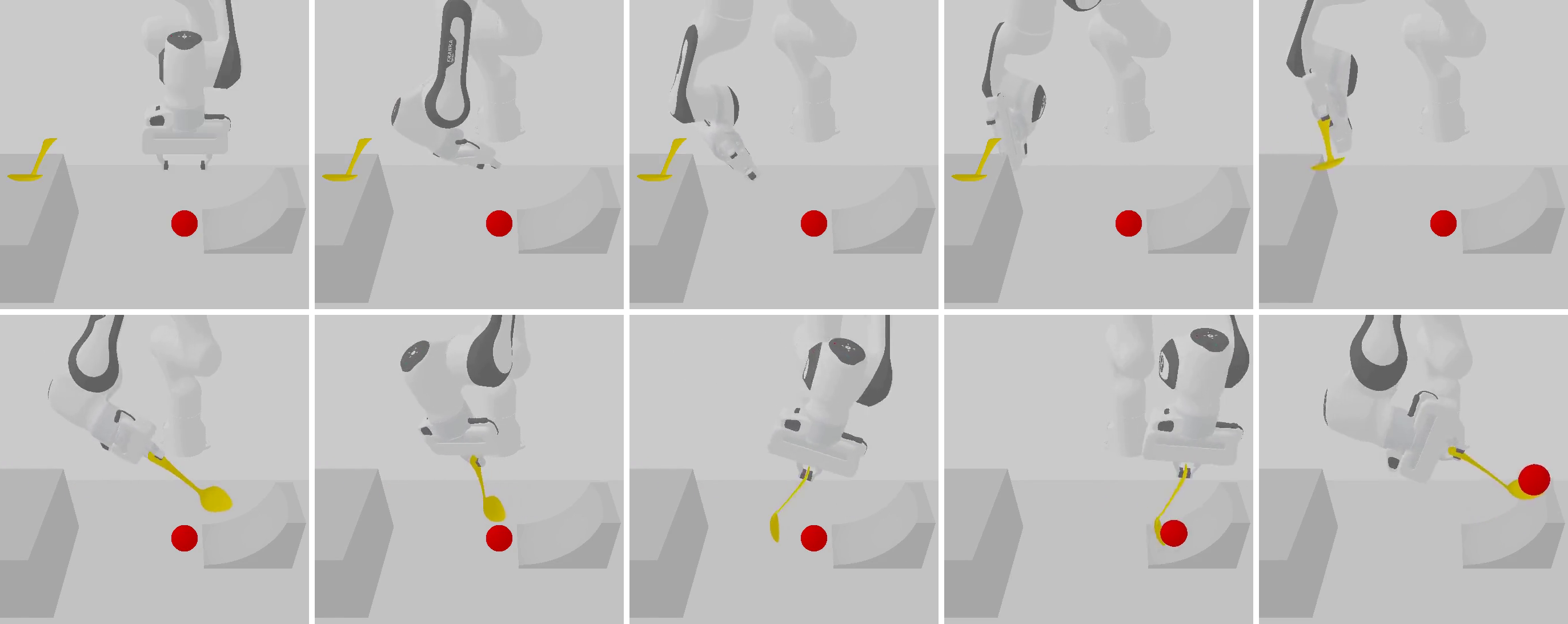}
    \caption{\textbf{Two example episodes of \textit{Scooping} in simulation.}}
    \label{fig:scooping_demo}
\end{figure}

\begin{figure}[htbp]
    \centering
    \includegraphics[width=\linewidth]{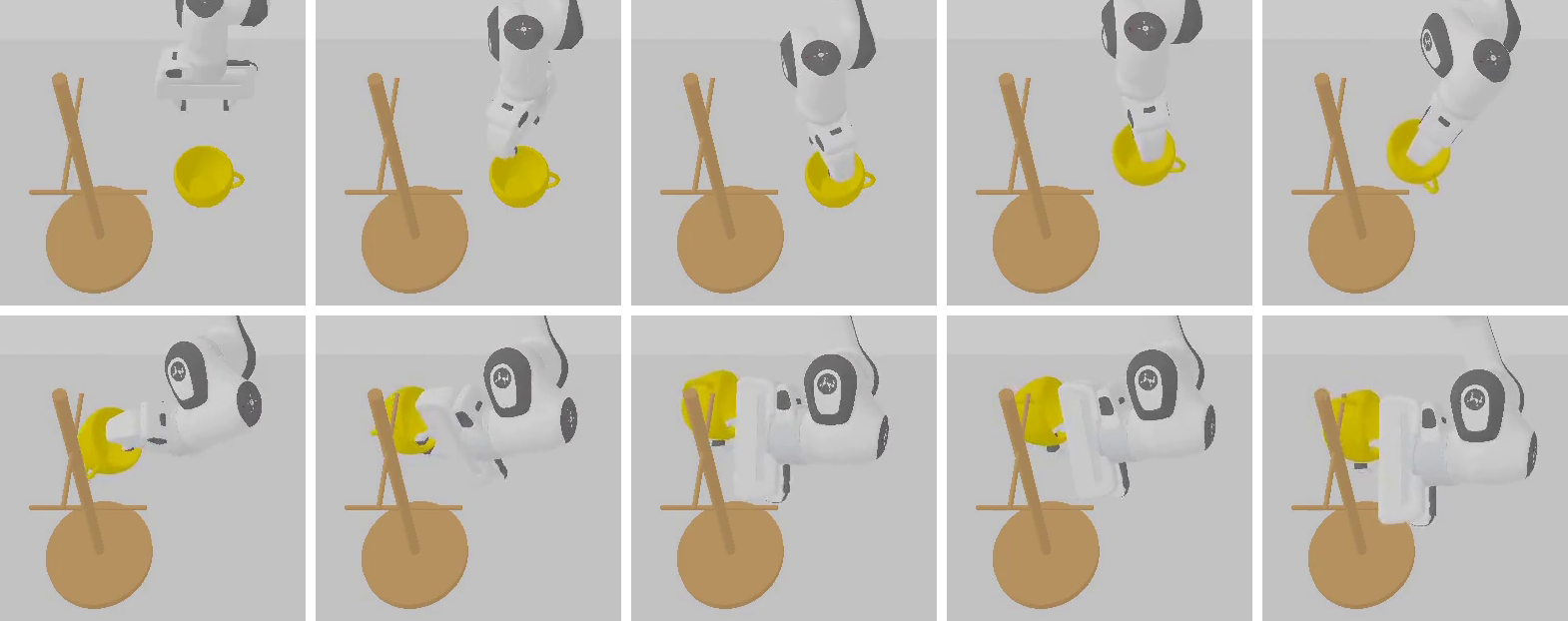}\\
    \vspace{0.05in}
    \includegraphics[width=\linewidth]{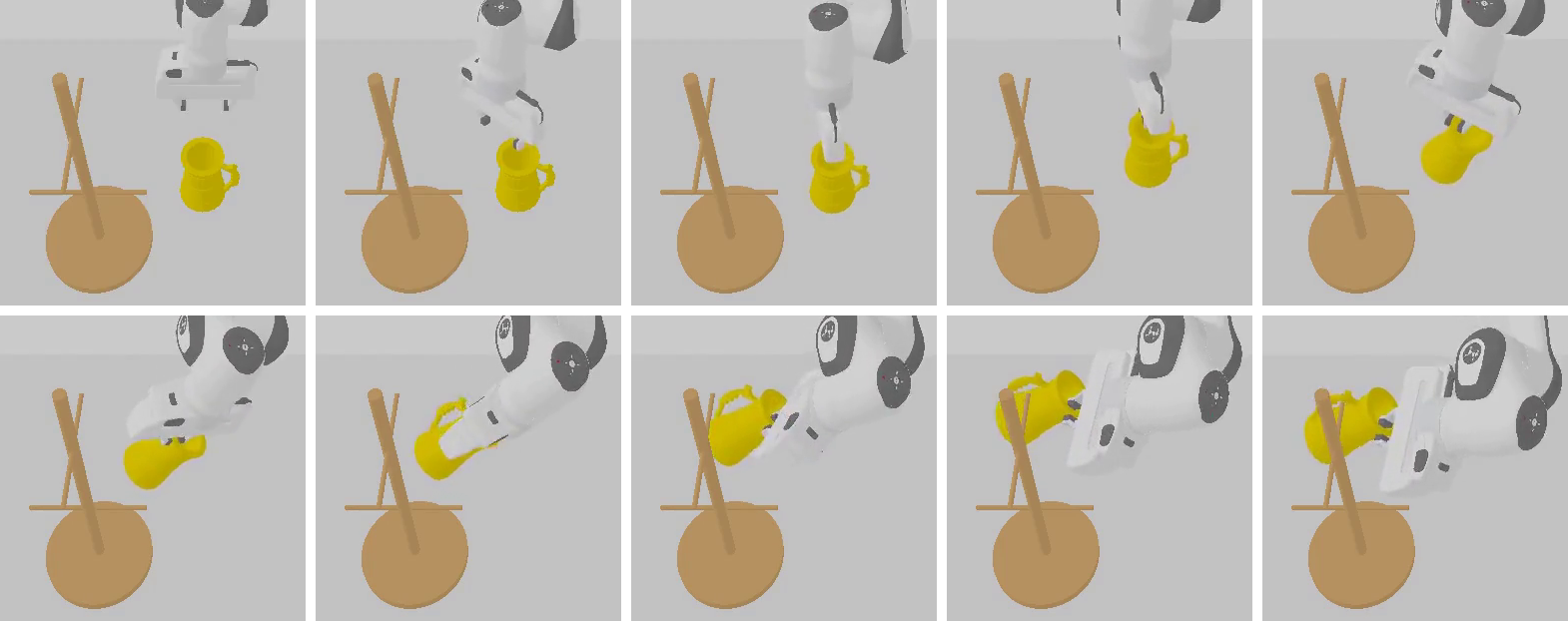}
    \caption{\textbf{Two example episodes of \textit{Hanging} in simulation.}}
    \label{fig:hanging_demo}
\end{figure}

\begin{figure}[htbp]
    \centering
    \includegraphics[width=\linewidth]{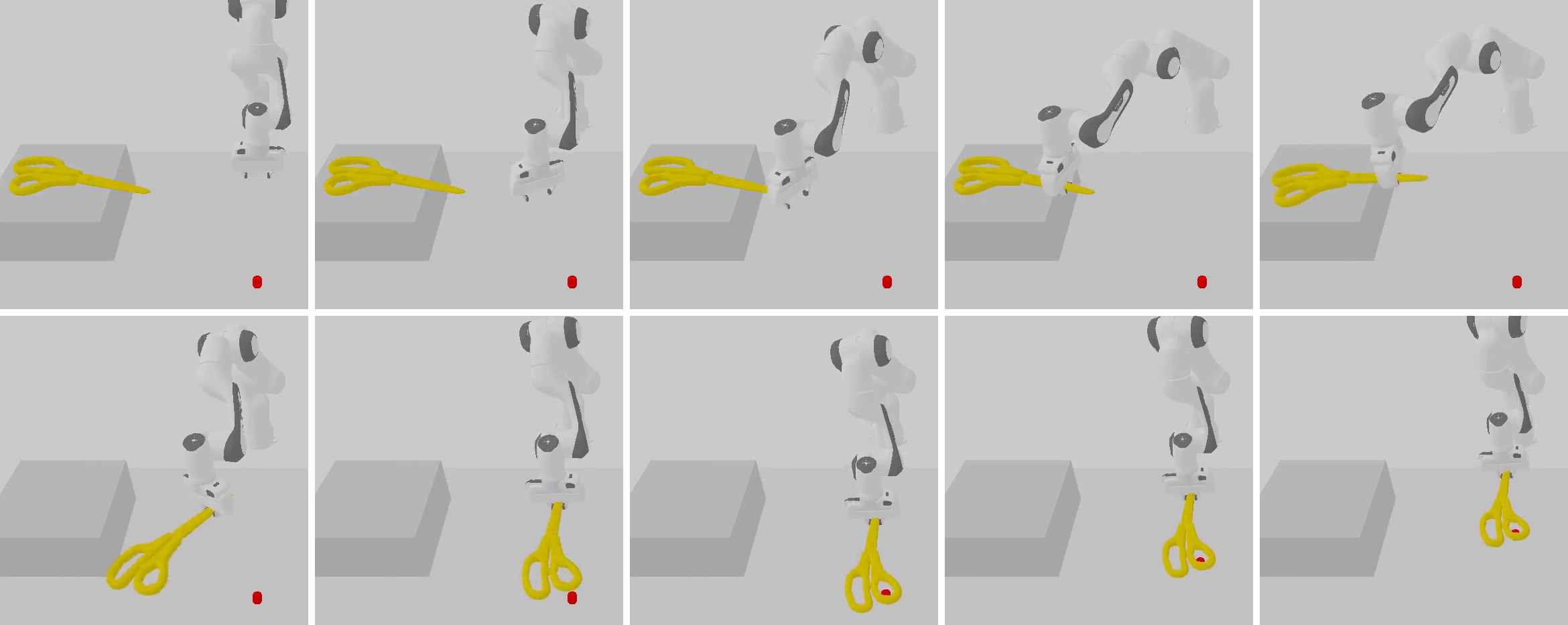}\\
    \vspace{0.05in}
    \includegraphics[width=\linewidth]{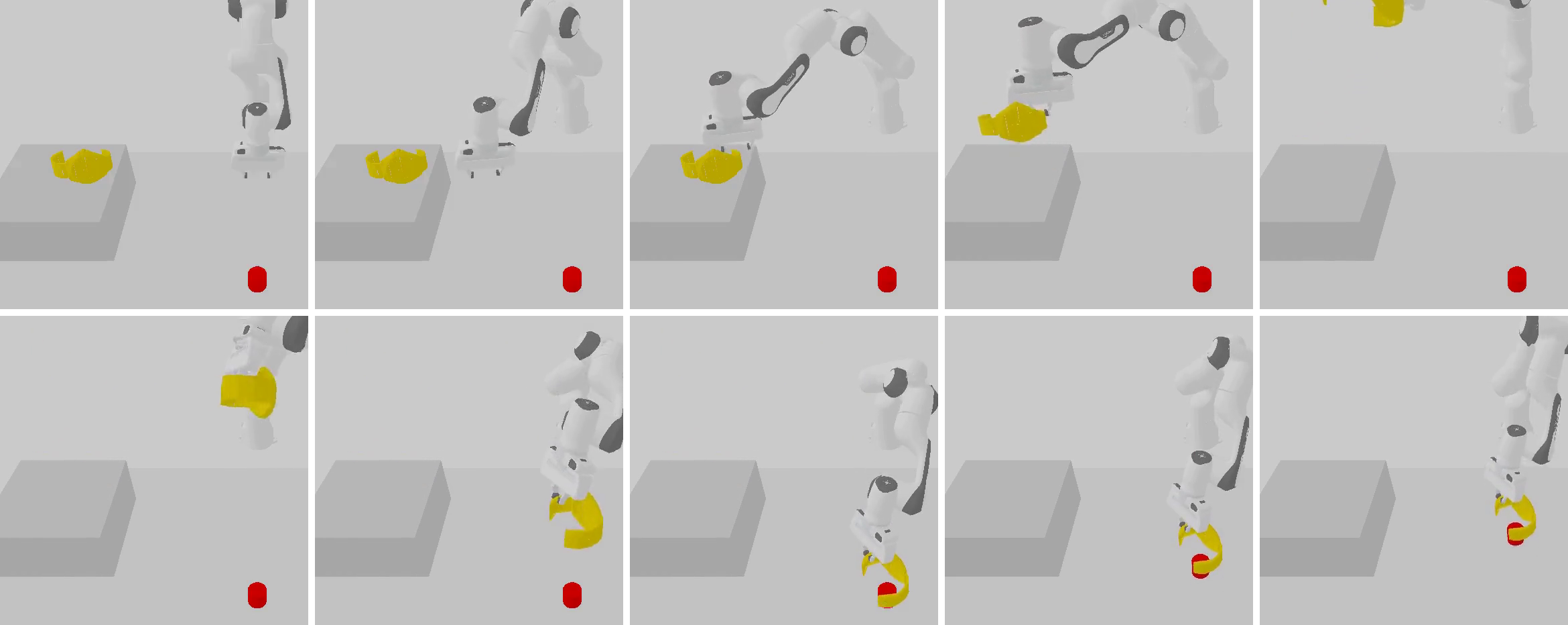}
    \caption{\textbf{Two example episodes of \textit{Hooking} in simulation.}}
    \label{fig:hooking_demo}
\end{figure}

\xhdr{Visualization of trajectories for simulation tasks.} In \fig{fig:scooping_demo},~\ref{fig:hanging_demo} and~\ref{fig:hooking_demo}, we provide two example trajectories of each task generated by \ours.

\xhdr{Visualization of trajectories for real-world tasks.} We demonstrate an example trajectory of \textit{Hanging} and \textit{Hooking} in \fig{fig:real_demo_combined}.

\begin{figure}
    \centering
    \includegraphics[width=\linewidth]{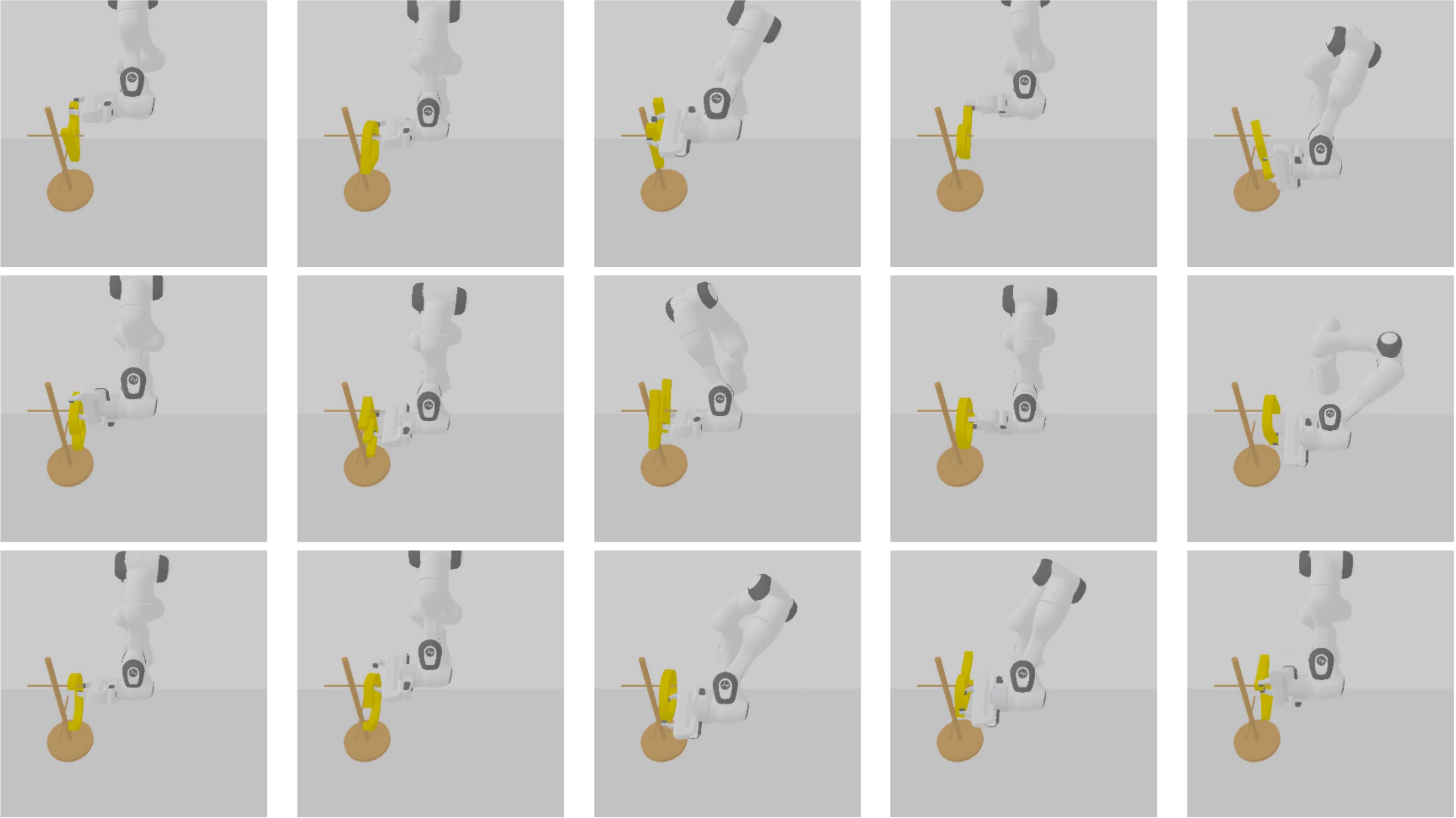}\\
    \vspace{0.05in}
    \includegraphics[width=\linewidth]{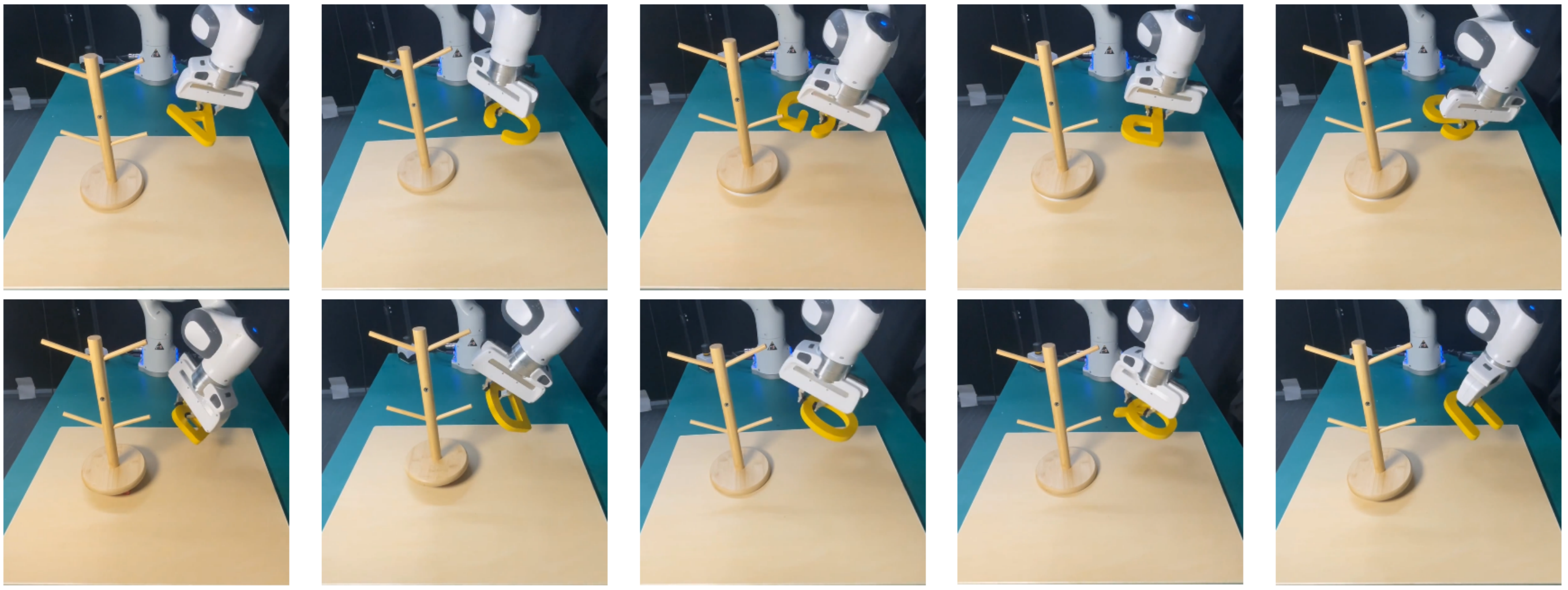}
    \caption{\textbf{Visualization of \textit{Hanging} alphabet objects in simulation (top) and real-world (bottom) settings.}}
    \vspace{-1.5em}
    \label{fig:hang_alphabet}
    \vspace{-1.6in}
\end{figure}

\begin{figure}
    \centering
    \includegraphics[width=\linewidth]{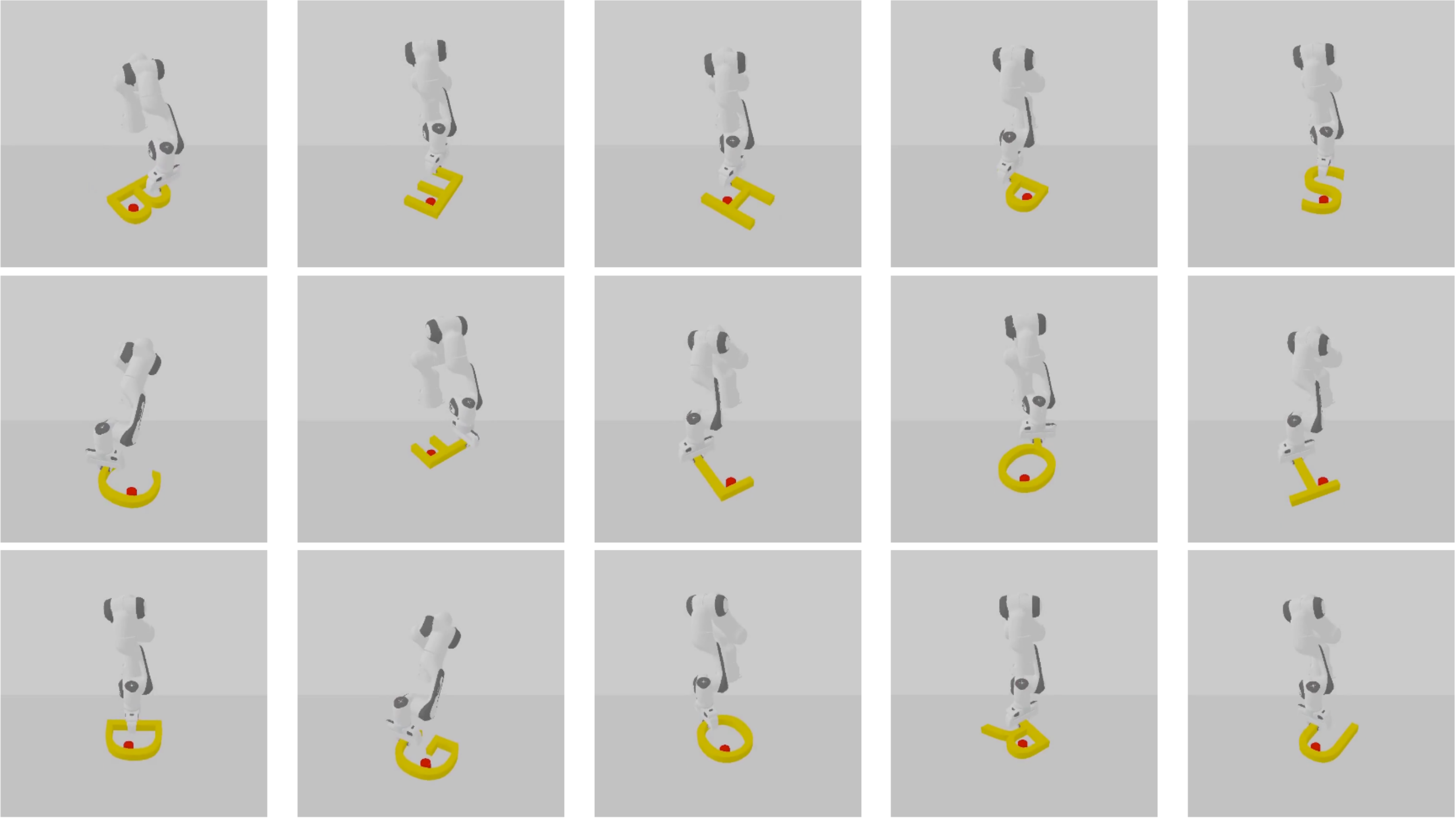}\\
    \vspace{0.05in}
    \includegraphics[width=\linewidth]{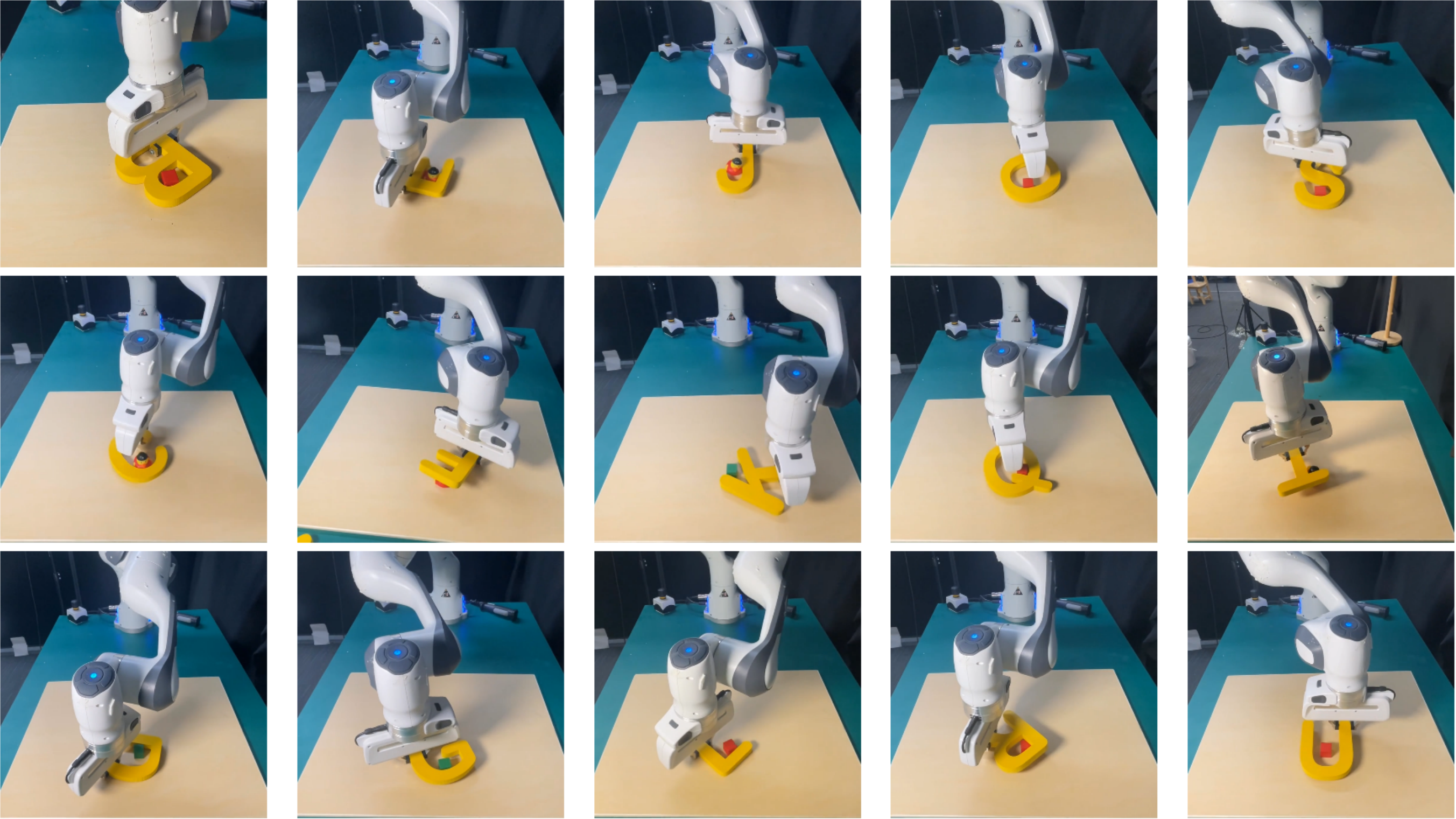}
    \caption{\textbf{Visualization of \textit{Hooking} with alphabet objects in simulation (top) and real-world (bottom) settings.}}
    \vspace{-1.5em}
    \label{fig:hook_alphabet}
\end{figure}

\xhdr{Additional Experiments on \textit{Hanging} and \textit{Hooking} with the \textit{Alphabet Toolkit}.} To further showcase our method's ability to generalize across objects of varying shapes, we conducted additional experiments on \textit{Hanging} and \textit{Hooking} tasks using the \textit{Alphabet Toolkit} in both simulation and real-world settings. Visualizations of these experiments are presented in Figures~\ref{fig:hang_alphabet} and~\ref{fig:hook_alphabet}. Please visit our project \href{https://magic-2024.github.io}{website} for the videos of the additional experiments.

\bibliographystyle{IEEEtran}
\bibliography{IEEEabrv,reference}

\end{document}